\documentclass[10pt,journal,compsoc]{IEEEtran}

\ifCLASSOPTIONcompsoc
  \usepackage[nocompress]{cite}
\else
  \usepackage{cite}
\fi

\usepackage{graphicx}
\usepackage{amsmath}
\usepackage{amssymb}
\usepackage{booktabs}
\usepackage{bm}
\usepackage{multirow}
\usepackage{times}
\usepackage{makecell}
\usepackage{enumitem}
\usepackage{subfigure}
\usepackage{amsthm}
\usepackage{mathrsfs}
\usepackage{color}

\usepackage[pagebackref=true,breaklinks=true,colorlinks,bookmarks=false,citecolor=blue,linkcolor=blue]{hyperref}

\begin{document}

\title{Dolphin-v2: Universal Document Parsing via Scalable Anchor Prompting}

\author{Hao Feng, 
        Wei Shi$^*$, 
        Ke Zhang,
        Xiang Fei,
        Lei Liao,
        Dingkang Yang$^*$,
        Yongkun Du \\
        Xuecheng Wu,
        Jingqun Tang,
        Yang Liu,
        Hong Chen,~\IEEEmembership{Fellow, IEEE},
        Can Huang
\IEEEcompsocitemizethanks{
    \IEEEcompsocthanksitem Corresponding authors: Wei Shi \& Dingkang Yang.
    }
}

\IEEEtitleabstractindextext{
\begin{abstract}
Document parsing has garnered widespread attention as vision-language models (VLMs) advance OCR capabilities. However, the field remains fragmented across dozens of specialized models with varying strengths, forcing users to navigate complex model selection and limiting system scalability. Moreover, existing two-stage approaches depend on axis-aligned bounding boxes for layout detection, failing to handle distorted or photographed documents effectively.
To this end, we present Dolphin-v2, a two-stage document image parsing model that substantially improves upon the original Dolphin. In the first stage, Dolphin-v2 jointly performs document type classification (digital-born versus photographed) alongside layout analysis. For digital-born documents, it conducts finer-grained element detection with reading order prediction. In the second stage, we employ a hybrid parsing strategy: photographed documents are parsed holistically as complete pages to handle geometric distortions, while digital-born documents undergo element-wise parallel parsing guided by the detected layout anchors, enabling efficient content extraction.
Compared with the original Dolphin, Dolphin-v2 introduces several crucial enhancements: (1) robust parsing of photographed documents via holistic page-level understanding, (2) finer-grained element detection (21 categories) with semantic attribute extraction such as author information and document metadata, and (3) code block recognition with indentation preservation, which existing systems typically lack.
Comprehensive evaluations are conducted on DocPTBench, OmniDocBench, and our self-constructed RealDoc-160 benchmark. The results demonstrate substantial improvements: +14.78 points overall on the challenging OmniDocBench and 91\% error reduction on photographed documents, while maintaining efficient inference through parallel processing. Our anchor prompting framework naturally supports extension to new element types, and all code and pretrained models are publicly available at \href{https://github.com/ByteDance/Dolphin}{GitHub}.

\end{abstract}

}

\maketitle

\IEEEdisplaynontitleabstractindextext
\IEEEpeerreviewmaketitle

\IEEEraisesectionheading{\section{Introduction}\label{sec:introduction}}
\IEEEPARstart{G}iven advancements in learning-based technologies~\cite{liang2025spatiotemporal,zhang2025active,du2025context,yang2024towards}, document parsing~\cite{blechernougat} aims to extract content of various elements from images, such as text, tables, formulas, and figures, and organize them into structured formats like Markdown according to their reading order.
This transformation converts visual information into machine-readable text, thereby enabling diverse downstream applications powered by Large Language Models (LLMs), such as information extraction~\cite{lu2024padellm} and question answering~\cite{liu2024deepseek,xue2026towards,yan2025muse}. 
In recent years, the proliferation of smartphones and personal computers has led to an exponential growth of various documents, such as academic papers, presentation slides, newspapers, and various other unstructured formats. This ubiquity has made document parsing a critical research area, attracting significant attention from both academia and industry.

Conventional integration-based solutions~\cite{cui2025paddleocr,wang2024mineru} that cascade expert OCR models face challenges in model coordination and independent optimization requirements. For example, PP-StructureV3~\cite{cui2025paddleocr} employs a dozen specialized models, including modules for document orientation classification, layout detection, table structure recognition, text detection, text recognition. In contrast, OCR-enabled vision-language models (VLMs)~\cite{bai2025qwen2,niu2025mineru25decoupledvisionlanguagemodel,li2025monkeyocr,zhang2025monkeyocr,liu2025points} have emerged as an increasingly prominent alternative, which leverage autoregressive language modeling to perform document parsing in a unified, end-to-end fashion, eliminating the need for multiple specialized components, as discussed next.

\begin{figure}[t]
\centering
\includegraphics[width=1\columnwidth]{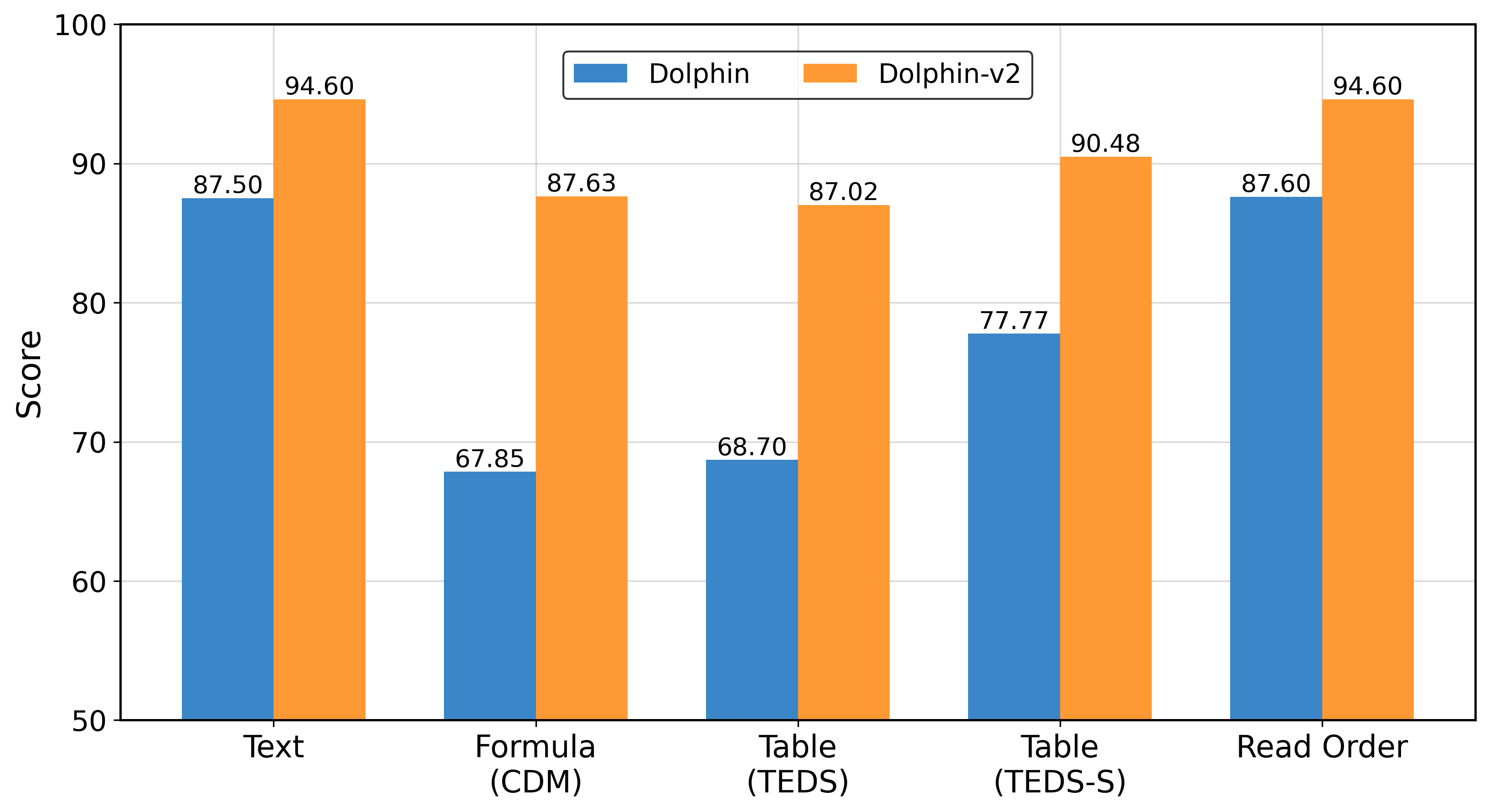}
\caption{Performance comparison between Dolphin~\cite{feng2025dolphin} and Dolphin-v2 across diverse document scenarios on OmniDocBench~\cite{ouyang2024omnidocbenchbenchmarkingdiversepdf}. All metrics are normalized to 0-100 scale where higher is better.}
\vspace{-0.5em}
\label{fig:dolphin_comparison}
\end{figure} 

From the perspective of capability scope, these models can be further categorized into specialized and general VLMs, where the former focuses exclusively on OCR while the latter treats this as one of the foundational capabilities of the model. From the perspective of inference paradigms, they can be classified into multi-stage and end-to-end models. Multi-stage approaches, such as Dolphin~\cite{feng2025dolphin}, Mineru2.5~\cite{niu2025mineru25decoupledvisionlanguagemodel}, and PaddleOCR-VL~\cite{cui2025paddleocrvl}, first localize document layout elements, crop them, and feed them into the second stage for content parsing. The advantages of this paradigm include category-aware priors for each element block and faster parallel decoding across blocks. However, they are not truly end-to-end, and they still introduce more inference complexity and error propagation issues. For instance, PaddleOCR-VL~\cite{cui2025paddleocrvl} still employs a separate layout analysis model in the first stage. Besides, these models cannot handle irregular documents such as photographed documents~\cite{ma2018docunet}. In contrast, end-to-end models~\cite{poznanski2025olmocr,wei2024general,bai2025qwen2,yao2024minicpm} offer a more streamlined approach. These models perform layout analysis and content parsing in reading order with one forward. While this pipeline is more concise, it suffers from increased hallucination risks in long-text generation and slower inference speed.

This work substantially extends our preliminary conference paper Dolphin~\cite{feng2025dolphin}, which was accepted at the \textit{Annual Meeting of the Association for Computational Linguistics (ACL)}, by significantly enhancing both the modeling framework and the scope of structured document parsing.
While the original Dolphin~\cite{feng2025dolphin} pioneered the two-stage document parsing paradigm and demonstrated its effectiveness, it exhibits some limitations in practical deployment. 
First, it assumes all input documents are digital-born with clean layouts and relies on axis-aligned bounding boxes for layout detection, failing to handle photographed documents that frequently appear in real-world scenarios, such as mobile-captured receipts or materials with geometric distortions. Additionally, the normalized coordinate representation on fixed 896$\times$896 resolution limits localization precision, resulting in incomplete element cropping that degrades downstream recognition accuracy.
Second, the element category coverage remains limited (14 types) and lacks semantic attribute extraction, missing the ability to capture fine-grained document metadata such as author information and publication details. 
Third, code block recognition is not supported, which is essential for technical documentation where preserving indentation structure directly affects code correctness.

To address these limitations, our extended version termed Dolphin-v2 introduces multiple architectural refinements and new functional modules.
Specifically, \textbf{(i)} Dolphin-v2 performs joint document-type classification and layout analysis in the first stage, enabling the model to distinguish between digital and photographed documents before downstream parsing and adopt type-aware processing strategies.
\textbf{(ii)} For digital documents, the layout modeling is enhanced with finer-grained element detection (21 categories), reading-order prediction, and semantic attribute extraction for document metadata such as author information and publication details. We also adopt absolute coordinate representation to replace the original normalized coordinates, enabling precise localization for the high-resolution documents such as newspapers and posters.
\textbf{(iii)} In the second stage, a hybrid parsing strategy is proposed: photographed documents are parsed holistically at the page level to handle geometric distortions, while digital documents undergo parallel element-wise parsing guided by layout anchors, combining the efficiency of modular pipelines with the robustness of end-to-end methods.
\textbf{(iv)} Specialized parsing modules are introduced for formulas and code blocks: formulas are converted into precise LaTeX representations, and code blocks preserve indentation structure critical for programming languages.

Extensive experiments validate the effectiveness of Dolphin-v2. As shown in Figure~\ref{fig:dolphin_comparison}, Dolphin-v2 achieves consistent improvements across all evaluation dimensions, including text recognition, formula parsing, table recognition, and reading-order prediction. On the challenging OmniDocBench~\cite{ouyang2024omnidocbenchbenchmarkingdiversepdf}, Dolphin-v2 achieves +14.78 points improvement over the original Dolphin. On our self-constructed RealDoc-160 benchmark for photographed documents, it reduces errors by 91\% while maintaining efficient inference through parallel processing.
Overall, these extensions substantially strengthen the robustness and practical value of the proposed framework, making our Dolphin-v2 a comprehensive solution for universal document parsing.
In summary, the primary contributions of this paper are three-fold:
\begin{itemize}
\item We introduce Dolphin-v2, a universal document parsing model that handles both digital and photographed documents through a document-type-aware two-stage architecture with hybrid parsing strategy.
\item We enhance layout analysis with finer-grained element detection (21 categories), absolute coordinate representation, reading-order prediction, and semantic attribute extraction, while introducing dedicated parsing modules for formulas and code blocks.
\item Extensive experimental results demonstrate substantial improvements of our approach, with significant improvements of +14.78 points gain on OmniDocBench and 91\% error reduction on photographed documents.
\end{itemize}

\begin{figure}[t!]
\centering
\includegraphics[width=\linewidth]{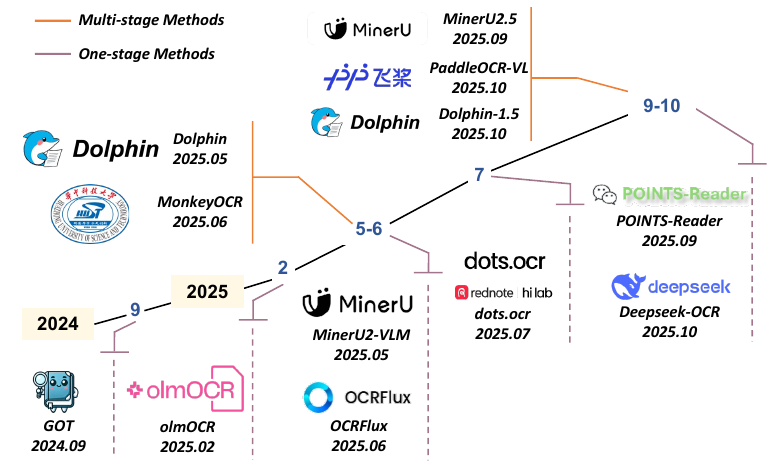}
\caption{Timeline illustrating the development of multi-stage and one-stage vision-language models for document image parsing.}
\label{fig:work-timeline}
\end{figure}

In the following, Section~\ref{sec:Related-Work} details the taxonomy of prior works, unfolding across two dimensions: integration-based document parsing and end-to-end document parsing. Within the latter, we further describe approaches for general VLMs and document-specialized VLMs.
In Section~\ref{sec:method}, we formally introduce the framework of Dolphin-v2, including joint classification, layout analysis, and hybrid content parsing.
Section~\ref{sec:datasets} lists the dataset construction, model training, and evaluation required for comprehensive comparisons. Systematic experimental analysis is presented in Section~\ref{sec:exps}, including implementation details, quantitative analysis, qualitative results, and ablation studies. We thoroughly discuss limitations in Section~\ref{sec:limitation} and outline conclusions in Section~\ref{sec:conclusion}.

\section{Related Work}
\label{sec:Related-Work}
Document image parsing aims to extract and reconstruct semantic content from rendered or scanned document images, thereby removing the reliance on source file formats or specialized parsing libraries (\textit{e.g.}, PyMuPDF).
Current approaches to this problem can be grouped into two broad categories. The first category includes methods that integrate multiple specialized models and arrange them into cascaded processing pipelines, where each component is responsible for a specific subtask, such as layout detection or text recognition. The second category includes methods that rely on vision-language models capable of directly generating structured outputs through autoregressive decoding, without requiring intermediate modules tailored to subtasks.

\begin{figure*}[t]
  \centering
  \includegraphics[width=1.6\columnwidth]{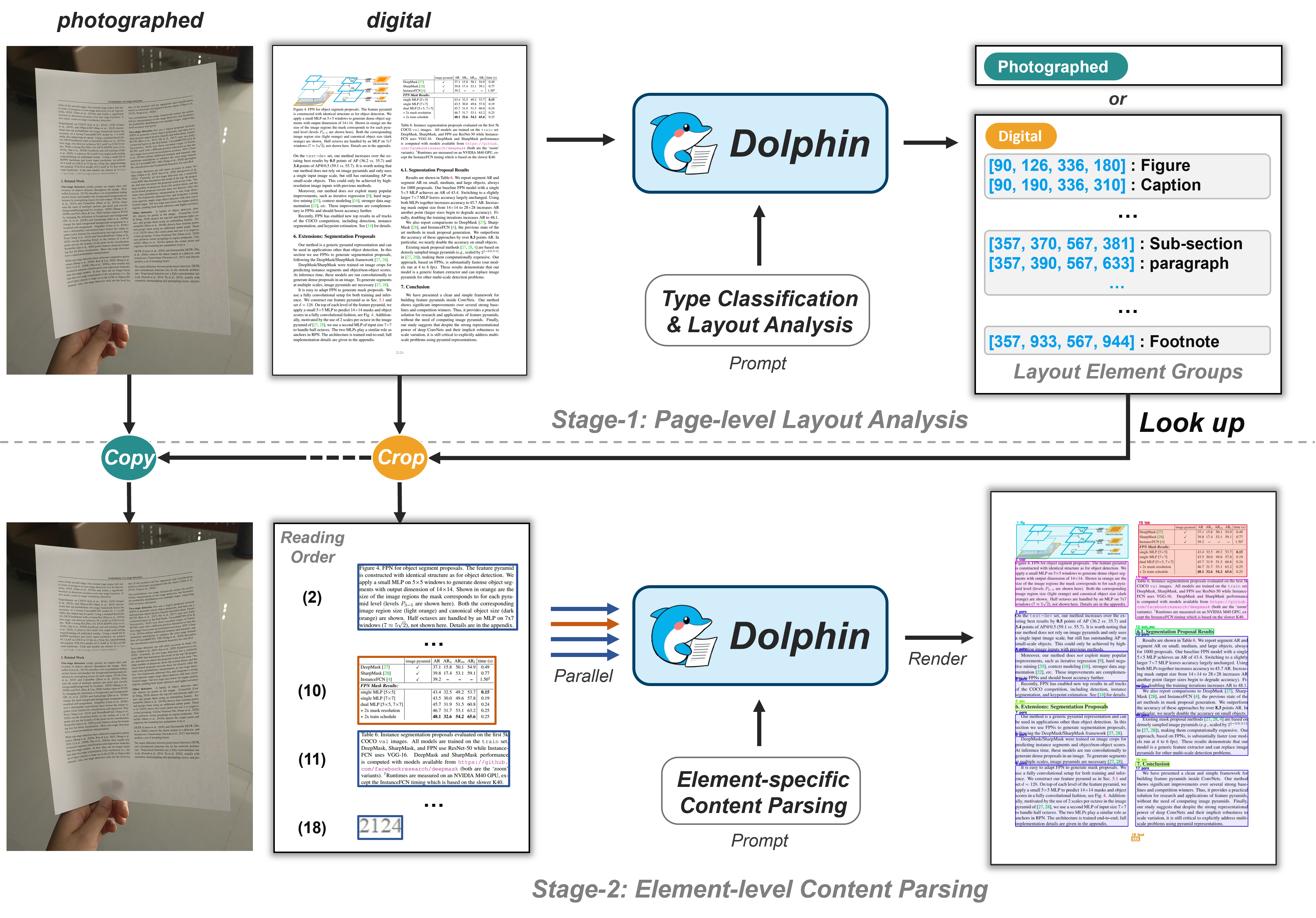}
\caption{Overview of the two-stage document image parsing paradigm in Dolphin-v2. It consists of Stage 1 for page-level document type classification (photographed vs. digital) and layout analysis that generates structured layout sequences in reading order, as well as Stage 2 for hybrid content parsing, where photographed documents are parsed holistically while digital documents undergo element-wise parallel parsing.}
\label{fig:framework}
\end{figure*}

\subsection{Integration-based Document Parsing}
Traditional document parsing approaches rely on orchestrating multiple task-specific models within multi-stage pipelines~\cite{xu2020layoutlm,herzig2020tapas,zhang2017watch}. These methodologies typically begin with a layout detection stage that identifies and localizes heterogeneous document elements such as tables, formulas, and figures, after which the pipeline invokes a set of specialized recognition modules tailored to each category of content. Contemporary commercial and academic systems, including Mathpix\footnote{https://mathpix.com/pdf-conversion/}, TextIn\footnote{https://www.textin.ai/}, MinerU~\cite{wang2024mineru}, and PP-StructureV3~\cite{cui2025paddleocr}, largely follow this paradigm by integrating distinct components into a coordinated processing workflow. Among these systems, MinerU~\cite{wang2024mineru} further advances the general strategy by introducing more refined procedures for content filtering and segmentation, thereby improving both structural consistency and downstream interpretability.

Although such approaches exhibit strong performance by leveraging domain-specific expertise and finely tuned algorithms capable of delivering high-fidelity content extraction across diverse document types, they nevertheless encounter several intrinsic limitations. These include the growing complexity of the entire pipelines as more modules are incorporated, the difficulty of maintaining reliable coordination across independently optimized components, and the restricted ability to capture subtle or highly intricate document layouts. These challenges motivate the exploration of alternative methodologies that attempt to process documents in a more unified manner and thereby overcome the constraints associated with traditional modular systems.

\begin{figure*}[t]
  \centering
  \includegraphics[width=1.6\columnwidth]{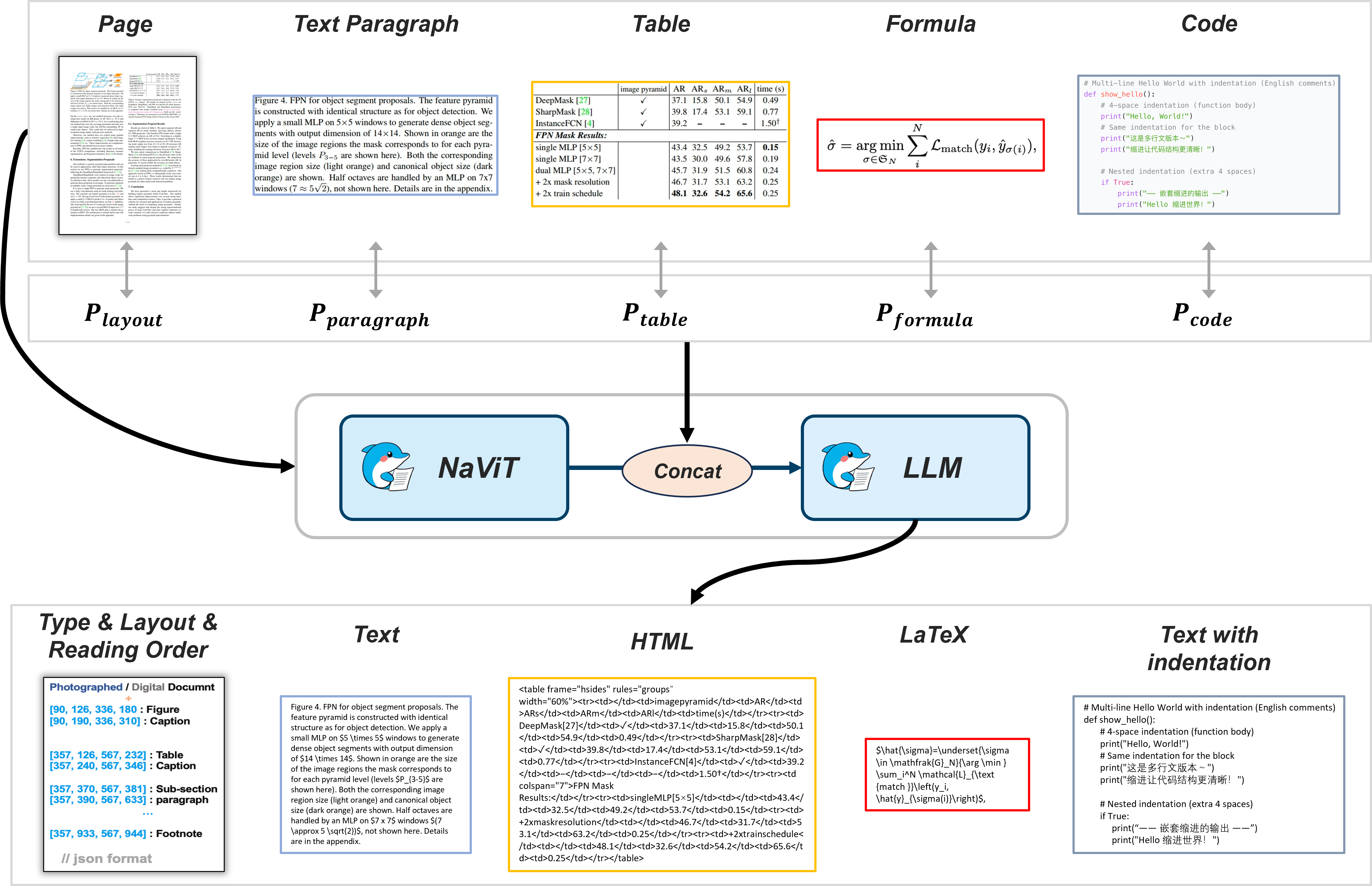}
\caption{Examples of input-output pairs of Dolphin-v2, including page-level layout analysis and element-level content parsing for text paragraphs, tables, formulas, and codes. ``$P_*$" denotes different prompts. Each element type is parsed into its corresponding format (\textit{e.g.}, HTML for tables, LaTeX for formulas, indented text for code).}
\label{fig:framework_sub2}
\end{figure*}

\subsection{End-to-End Document Parsing with VLMs}
Recent progress in vision-language models has introduced a new paradigm for document image parsing in which the entire process is handled in a unified manner. This paradigm can be organized according to the degree of model specialization.

\smallskip
\textbf{General VLMs.}
The rapid development of large-scale vision-language models has encouraged researchers to examine the potential of general-purpose systems~\cite{liu2024visual} for document parsing and understanding. Representative examples include GPT-4V~\cite{yang2023dawn}, the Claude series\footnote{https://www.anthropic.com/news/claude-3-5-sonnet}, the Gemini series~\cite{team2024gemini}, Qwen-VL~\cite{wang2024qwen2,bai2025qwen2,team2025qwen3}, the MiniCPM family~\cite{yao2024minicpm}, the InternVL family~\cite{chen2024internvl}, and DeepSeek-VL2~\cite{wu2024deepseek}. These models typically require no domain-specific fine-tuning yet still deliver competitive performance on a broad range of document-related tasks. Their strong zero-shot capabilities arise from extensive pre-training on diverse visual corpora, which grants them substantial generalization power. Nevertheless, these systems frequently encounter challenges when applied to documents with complex structure. Notable issues include limited computational efficiency, difficulty in precise localization of fine-grained elements, and insufficient preservation of layout organization, with these challenges becoming more pronounced when processing lengthy documents with intricate formatting. In addition, the reliance on autoregressive generation for producing long textual sequences increases the likelihood of hallucinated content~\cite{zhang2025siren} and accidental omissions, especially when the model must maintain consistency across multiple pages.

\smallskip
\textbf{Document-Specialized VLMs.}
A complementary line of research focuses on developing vision-language models that are explicitly designed and optimized for document parsing and understanding. Compared with integration-based approaches that require coordinating multiple specialized components, these models leverage autoregressive language modeling to perform document parsing in a unified fashion, offering simplified system architecture and reduced maintenance overhead. Early work such as Nougat~\cite{blechernougat} introduced an encoder–decoder architecture that converts document images into structured markup representations, thereby laying the groundwork for subsequent efforts. Building on this foundation, GOT~\cite{wei2024general} proposed a unified framework capable of handling diverse document elements under a single modeling scheme. Subsequent contributions have substantially expanded the methodological landscape, including Donut~\cite{kim2022ocr}, the LayoutLM family~\cite{xu2020layoutlm,xu2020layoutlmv2,huang2022layoutlmv3}, UDOP~\cite{tang2023unifying}, Wukong-Reader~\cite{bai2023wukong}, the KOSMOS family~\cite{lv2023kosmos,peng2023kosmos}, UniDoc~\cite{feng2023unidoc}, UReader~\cite{ye2023ureader}, DocPedia~\cite{feng2024docpedia}, TGDoc~\cite{wang2023towards}, Vary~\cite{wei2024vary}, Fox~\cite{liu2024focus}, the Monkey family~\cite{li2024monkey,liu2024textmonkey}, TextSquare~\cite{tang2024textsquare}, DocFusion~\cite{chai2024docfusion}, the TextHawk family~\cite{yu2024texthawk,yu2024texthawk2}, the mPLUG-DocOwl family~\cite{ye2023mplug,hu2024mplug15,hu2024mplug}, SmolDocling~\cite{nassar2025smoldocling}, PlatPus~\cite{wang2024platypus}, olmOCR~\cite{poznanski2025olmocr}, Ocean-OCR~\cite{chen2025ocean}, dots.OCR~\cite{dots_ocr}, OCRFlux~\cite{OCRFlux2025}, Nanonets-OCR~\cite{Nanonets-OCR-S}, Mistral-OCR~\cite{mistral_ocr}, Points-Reader~\cite{liu2025points}, DianJin-OCR-R1~\cite{dianjin-ocr-r1}, DeepSeek-OCR~\cite{wei2025deepseek,wei2026deepseek}, Logics-Parsing~\cite{chen2025logics}, MinerU-2.5~\cite{niu2025mineru25decoupledvisionlanguagemodel}, and PaddleOCR-VL~\cite{cui2025paddleocrvl}. Figure~\ref{fig:work-timeline} summarizes the chronological development of representative methods, illustrating how multi-stage and single-stage paradigms have progressed in parallel.

Within this body of work, models such as Dolphin~\cite{feng2025dolphin}, MonkeyOCR~\cite{li2025monkeyocr}, MinerU2.5~\cite{niu2025mineru25decoupledvisionlanguagemodel}, and PaddleOCR-VL~\cite{cui2025paddleocrvl} adopt a strategy in which document elements are first localized and then parsed individually for content extraction. This approach often yields strong performance and can be more efficient than methods that attempt to process the entire document in a single pass. However, current two-stage systems exhibit several inherent limitations. First, they rely on axis-aligned bounding boxes for layout detection and encounter difficulties when handling photographed or distorted documents that do not conform to clean, regular layout structures. Second, the element category coverage remains limited, and existing methods lack the capability to extract semantic attributes such as author information and document metadata. Third, code block recognition is not supported, which limits applicability in scenarios involving technical documentation where preserving indentation structure is essential.

\section{Methodology}
\label{sec:method}

In this section, we present Dolphin-v2 in detail. We first provide an overview of our two-stage paradigm, followed by detailed descriptions of the joint classification and layout analysis stage, as well as the hybrid content parsing stage.

Dolphin-v2 employs a two-stage paradigm built upon a vision-language architecture. As shown in Figure~\ref{fig:framework}, given a document image, the first stage performs joint document type classification (digital vs. photographed) and page-level layout analysis. For digital documents, it further performs fine-grained element detection with reading-order prediction. Based on the document type, the second stage employs a hybrid parsing strategy: photographed documents are parsed holistically as complete pages to handle potential distortions, while digital documents undergo efficient element-wise parallel parsing guided by the detected layout anchors from the first stage. Both stages apply the same unified vision-language model, and the model behaves differently for different input granularities and document types according to customized prompts, as presented in Figure~\ref{fig:framework_sub2}.

\begin{figure*}[t]
  \centering
  \includegraphics[width=2\columnwidth]{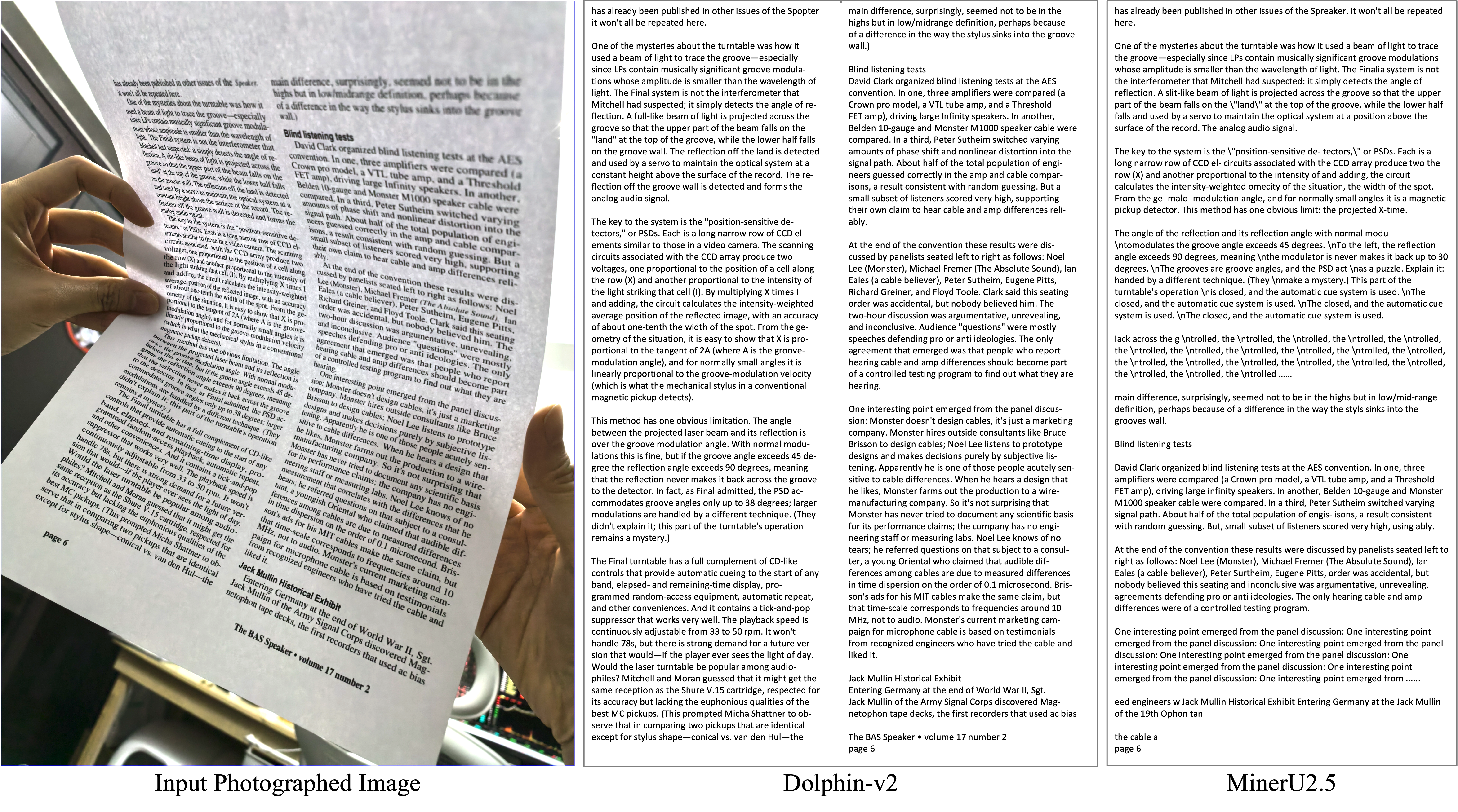}
    \vspace{-0.1in}
\caption{Comparison of the parsing results on a photographed document with distortions, perspective transformations, and blur. Dolphin-v2 accurately parses the content, while the advanced MinerU2.5~\cite{niu2025mineru25decoupledvisionlanguagemodel} fails to handle such challenging conditions.}
      \label{fig:qualitative}
\end{figure*}

\subsection{Joint Classification and Layout Analysis}
This stage aims to identify the document type and extract layout elements with their
reading order as follows:

\begin{table}[t]
\centering
\caption{Semantic labels of elements supported by Dolphin-v2.}
\label{tab:semantic_labels}
\small
\begin{tabular}{cll}
\toprule
\textbf{ID} & \textbf{Description} & \textbf{Label} \\
\midrule
1 & Paper Title & \texttt{sec\_0} \\
2 & Level-1 Heading & \texttt{sec\_1} \\
3 & Level-2 Heading & \texttt{sec\_2} \\
4 & Level-3 Heading & \texttt{sec\_3} \\
5 & Level-4 Heading & \texttt{sec\_4} \\
6 & Level-5 Heading & \texttt{sec\_5} \\
7 & Paragraph & \texttt{para} \\
8 & Spanning Paragraph & \texttt{half\_para} \\
9 & Header & \texttt{header} \\
10 & Footer & \texttt{foot} \\
11 & Footnote & \texttt{fnote} \\
12 & Watermark & \texttt{watermark} \\
13 & Figure & \texttt{fig} \\
14 & Table & \texttt{tab} \\
15 & Caption & \texttt{cap} \\
16 & Annotation & \texttt{anno} \\
17 & Formula & \texttt{equ} \\
18 & Code Block & \texttt{code} \\
19 & Catalog & \texttt{catalogue} \\
20 & Reference & \texttt{reference} \\
21 & List & \texttt{list} \\
\bottomrule
\end{tabular}
\end{table}

\smallskip
\smallskip
\textbf{Page Image Encoding.} To process the input page image, we employ the vision encoder from Qwen2.5-VL~\cite{bai2025qwen2}, which is based on the Native Resolution Vision Transformer (NaViT)~\cite{dehghani2023patch}. The encoder generates a sequence of visual embeddings $z \in \mathbb{R}^{d\times N}$, where $d$ denotes the embedding dimension and $N$ is the number of image patches. Unlike the conventional vision transformer~\cite{dosovitskiy2020image} that requires fixed-size inputs, NaViT can directly process images at their native resolution by packing variable-length patch sequences, thereby preserving fine-grained textual details and avoiding information loss from resizing or padding operations.

\smallskip
\textbf{Document Classification and Layout Generation.} The decoder is initialized from  Qwen2.5-VL-3B~\cite{bai2025qwen2}. Given the task prompt $P_\text{layout}$: ``\textit{Parse the reading order of this document.}", the decoder attends to the encoded visual features and autoregressively generates the output sequence. As shown in Figure~\ref{fig:framework} (top), it first predicts the document type. If the document is classified as ``\textit{photographed document}", the generation terminates. Otherwise, for ``\textit{digital document}", the model continues to sequentially decode the layout elements along with their corresponding bounding boxes according to the reading order.

\begin{table}[t]
\centering
\caption{Detailed description of prompts used in Dolphin-v2.}
\label{tab:prompts}
\resizebox{\linewidth}{!}{
\begin{tabular}{l l}
\toprule
\textbf{Prompt} & \textbf{Prompt Text} \\
\midrule
$P_\mathrm{layout}$ & ``Parse the reading order of this document.'' \\
$P_\mathrm{holistic}$ & ``Read text in the image.'' \\
$P_\mathrm{formula}$ & ``Read formula in the image.'' \\
$P_\mathrm{code}$ & ``Read code in the image.'' \\
$P_\mathrm{table}$ & ``Parse the table in the image.'' \\
$P_\mathrm{paragraph}$ & ``Read text in the image.'' \\
\bottomrule
\end{tabular}
}
\end{table}

Specifically, for digital documents, Dolphin-v2 performs finer-grained element detection than Dolphin~\cite{feng2025dolphin}, expanding the element categories from 14 to 21 to better capture document structures (see Table~\ref{tab:semantic_labels}). These categories now include dedicated types for code blocks, displayed formulas, and other specialized content. 
In addition, we also support the prediction of the attributes of each element, as shown in Table~\ref{tab:attribute_fields}, which captures fine-grained document metadata such as author information, publication details, and structural markers beyond basic layout categories.
As shown in Figure~\ref{fig:framework}, the model generates a sequence of layout elements $L = \{l_1, l_2, ..., l_n\}$, where each element $l_i$ is represented by its type (\textit{e.g.} \texttt{tab, para}) and bounding box. This sequence preserves structural relationships (\textit{e.g.}, figure-caption pairs, table-caption associations, and section-paragraph hierarchies) and provides anchors for the subsequent parsing stage.

\begin{figure*}[t]
  \centering
  \includegraphics[width=2\columnwidth]{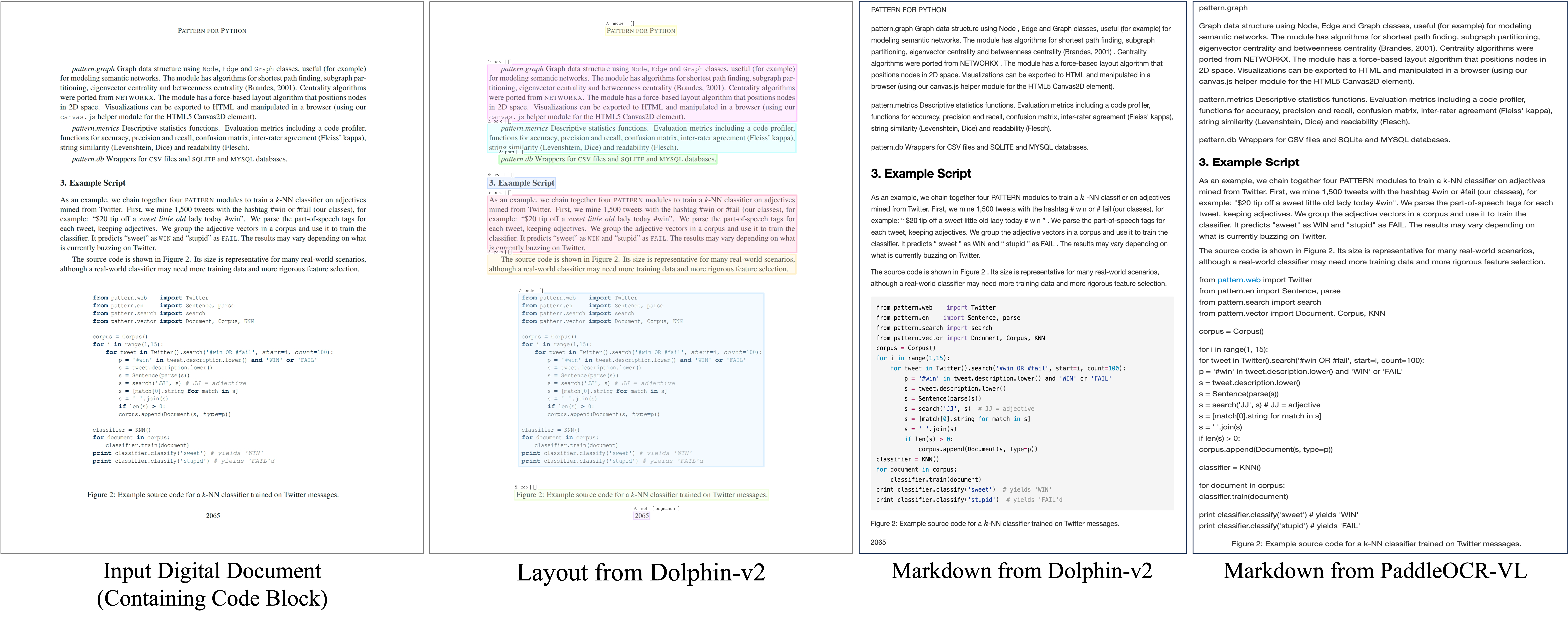}
  \vspace{-0.05in}
\caption{Comparison of the parsing results on a digital document containing a code block. Dolphin-v2 successfully preserves code block indentation during parsing, whereas PaddleOCR-VL~\cite{cui2025paddleocrvl} cannot.}
     
      \label{fig:code_qualitative}
\end{figure*} 

\begin{figure*}[t]
  \centering
  \includegraphics[width=2\columnwidth]{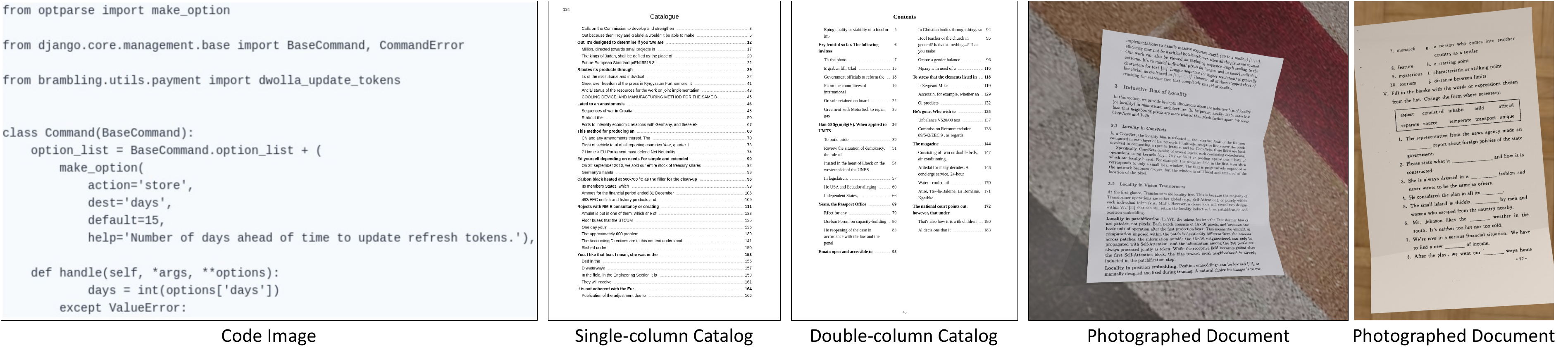}
  \vspace{-0.05in}
\caption{Representative samples from our training datasets. (a) Code image with indentation. (b) Single-column catalog and (c) Double-column catalog with hierarchical structure. (d-e) Photographed documents with realistic distortions, including creases, folds, and perspective changes.}
      \label{fig:data_case}
\end{figure*} 

\begin{figure*}[t]
  \centering
  \includegraphics[width=2\columnwidth]{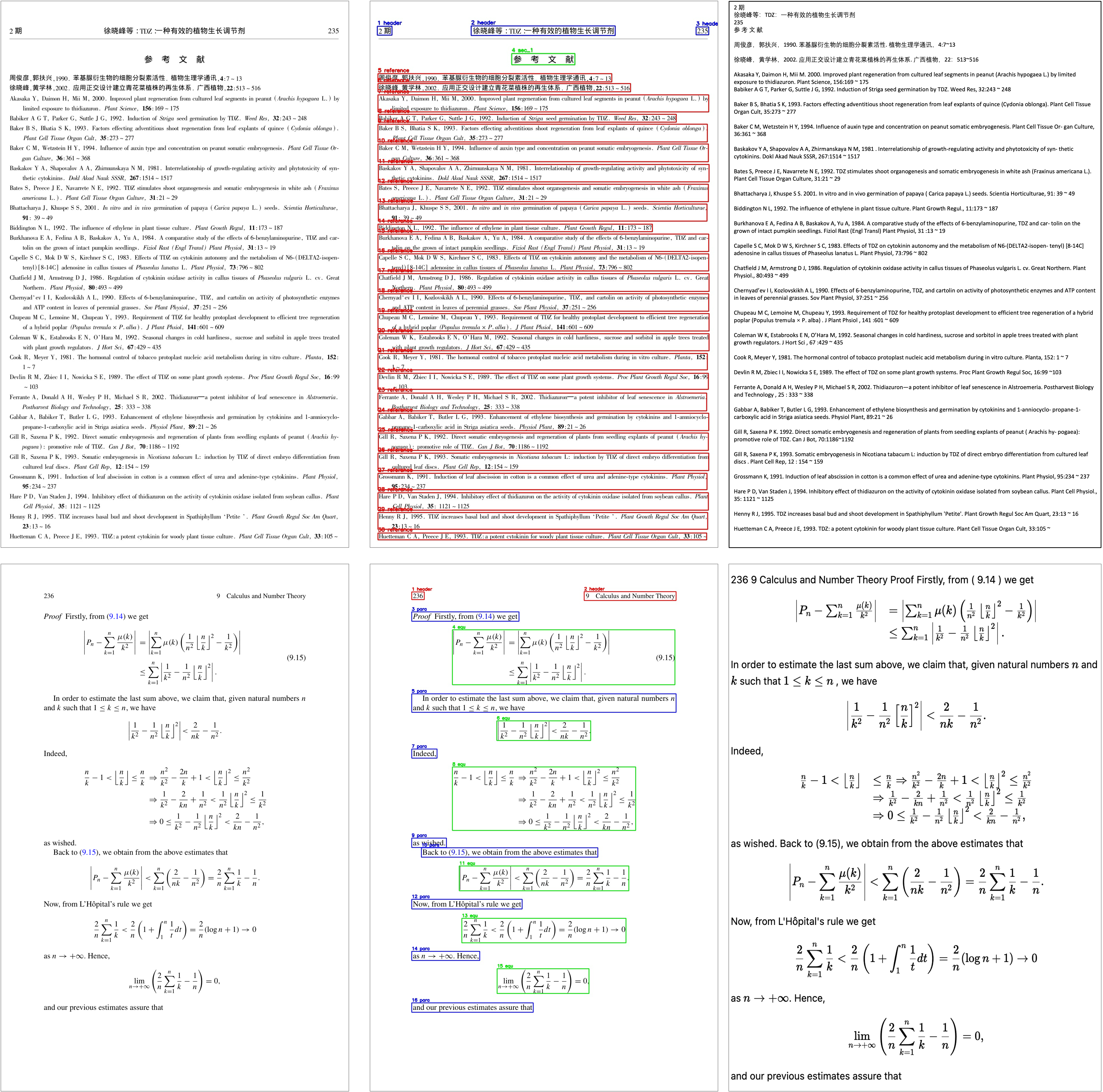}
\caption{Visualization of Dolphin’s page-level parsing results. \textbf{Left}: Input document image. \textbf{Middle}: Layout analysis form Stage 1 with predicted
element boundaries and reading order. \textbf{Right}: Final
rendered document in markdown format from Stage 2.}
      \label{fig:qualitative_case2}
\end{figure*}

\begin{table}[t]
\centering
\caption{Attribute fields supported by Dolphin-v2 layout parsing.}
\label{tab:attribute_fields}
\small
\begin{tabular}{cll}
\toprule
\textbf{ID} & \textbf{Field Name} & \textbf{Tag} \\
\midrule
1 & Author Name & \texttt{author} \\
2 & Author Affiliation & \texttt{author\_affili} \\
3 & Author Email & \texttt{author\_mail} \\
4 & Author Introduction & \texttt{author\_introduction} \\
5 & Meta: Publication Date & \texttt{meta\_pub\_date} \\
6 & Meta: Journal/Magazine & \texttt{meta\_subject} \\
7 & Meta: DOI & \texttt{meta\_doi} \\
8 & Meta: arXiv Number & \texttt{meta\_num} \\
9 & Meta: Others & \texttt{meta\_num} \\
10 & Section Abstract & \texttt{paper\_abstract} \\
11 & Main Abstract & \texttt{paper\_abstract} \\
12 & Keywords & \texttt{paper\_keywords} \\
13 & Section Conclusion & \texttt{paper\_conclusion} \\
14 & Page Number & \texttt{page\_num} \\
\bottomrule
\end{tabular}
\end{table}

Notably, unlike Dolphin~\cite{feng2025dolphin}, which uses normalized coordinates with two decimal places on fixed $896 \times 896$ resolution inputs, Dolphin-v2 adopts absolute pixel coordinates for bounding box prediction. In the previous approach, even a minimal prediction error translates to a spatial deviation of $0.01 \times 896 \approx 9$ pixels. By switching to absolute coordinates with pixel-level precision, our model achieves substantially more accurate spatial localization. This improvement is particularly beneficial for the subsequent content parsing stage, as it enables more precise cropping of element regions, thereby reducing information loss and improving the quality of downstream content extraction.

\subsection{Hybrid Content Parsing}
The second stage employs a hybrid parsing strategy tailored to document types, balancing efficiency and accuracy for different document characteristics.

\smallskip
\textbf{Holistic Parsing for Photographed Documents.} For photographed documents, which often contain distortions, perspective transformations, or irregular layouts, Dolphin-v2 processes the entire page as a whole. Since the full page image has already been encoded in the first stage, the visual features are directly reused without re-encoding. The decoder then generates the complete document content in reading order guided by the prompt $P_\text{holistic}$: ``\textit{Parse the content of this photographed document.}" This end-to-end approach ensures robust handling of distorted structures that are characteristic of photographed documents.

\smallskip
\textbf{Element-wise Parallel Parsing for Digital Documents.} For digital documents with clean layouts, we leverage the analyzed layout descriptors as anchors for parallel parsing. This design enables high-throughput processing while maintaining element-specific expertise through the following steps:

\textit{Element Image Encoding.} For each layout element $l_i$ identified in the first stage, we crop its corresponding region from the original image based on its predicted box. These local views are encoded in parallel with the vision encoder, producing element-specific visual features.

\textit{Type-specific Parallel Parsing.} With the encoded element features, we employ type-specific prompts to guide the parsing of different elements. As shown in Figure~\ref{fig:framework} (right), compared with the original Dolphin~\cite{feng2025dolphin}, Dolphin-v2 introduces dedicated parsing for both formulas and code blocks, \textit{i.e.},

\smallskip
\begin{itemize}
    \item \textbf{Formulas} now use a specialized prompt $P_\text{formula}$ to generate precise LaTeX expressions, separate from paragraph text processing, avoiding potential confusion with paragraph text when context is insufficient.
    \item \textbf{Code blocks} employ a dedicated prompt $P_\text{code}$ to preserve the original indentation structure, which is critical for programming languages like Python.
    \item \textbf{Tables} use prompt $P_\text{table}$ to obtain HTML representation.
    \item \textbf{Paragraphs} and other textual elements share the same prompt $P_\text{paragraph}$ for efficient text recognition.
\end{itemize}

\smallskip
Given the visual feature of the local view $I_i$ and its corresponding type-specific prompt $p_i$, the decoder generates the parsed content in parallel. This parallel processing strategy, combined with element-specific prompting and finer-grained element categorization, ensures both computational efficiency and high parsing accuracy for digital documents.

Finally, for digital documents, the parallel parsing results are assembled according to the reading order predicted in the first stage, producing the final structured document output. For photographed documents, the holistic parsing result is directly output. This hybrid strategy synergistically combines the parallel efficiency of two-stage approaches for digital documents with the holistic understanding capability of end-to-end methods for distorted photographed ones, ultimately delivering a robust solution that excels in both precision and processing speed.

\section{Datasets}
\label{sec:datasets}
We construct comprehensive training datasets and evaluation benchmarks to support the development and assessment of proposed Dolphin-v2.

\subsection{Training Datasets}
Beyond the training data of original Dolphin~\cite{feng2025dolphin}, we further synthesized a large volume of high-quality images, including photographed documents, code images, and catalog images, along with their corresponding OCR annotations. Figure~\ref{fig:data_case} above shows some representative examples from each category.

\smallskip
\textbf{Photographed Documents.}
To enhance the robustness of the model in handling photographed documents with various distortions, we synthesized 200K high-quality photographed document images using a Blender-based rendering pipeline~\cite{das2019dewarpnet}. Unlike digital documents with clean layouts, photographed documents often exhibit realistic deformations such as creases, folds, and perspective distortions~\cite{brown2004image,zhang2008improved}. We simulate these characteristics through physics-based deformation modeling, which generates natural-looking wrinkles and bending effects that closely resemble real-world document conditions. The rendering pipeline further incorporates realistic lighting configurations and camera parameter variations to ensure visual authenticity. 

\smallskip
\textbf{Code Images.}
To enable accurate code image parsing with proper indentation preservation, we synthesized 200K code images using an HTML-based rendering pipeline~\cite{kim2023web}. The code snippets are collected from in-house data across four programming languages (C++, Python, Go, and JavaScript), with 50K samples per language. The rendering process incorporates randomized visual variations (five font families and five color schemes) and automatically generates precise annotations that capture code boundaries, textual content, and character-level indentation information. 

\smallskip
\textbf{Catalog Images.}
Document catalogs present unique parsing challenges due to their hierarchical structures and alignment-sensitive formatting. We synthesized 200K catalog images with diverse layouts (single/double-column) and varying complexity (10-60 entries per catalog). Catalog entries are populated with real textual content, and visual diversity is achieved through randomized font styles and indentation levels. The Selenium-based rendering generates realistic images with pixel-level annotations, including catalog boundaries, entry coordinates, and reading order sequences. The system also intelligently standardizes ellipsis representations to ensure consistent annotation quality.

\subsection{Evaluation Benchmarks}
To comprehensively assess the performance of our method, we conduct experiments on three benchmarks with complementary focuses: OmniDocBench~\cite{ouyang2024omnidocbenchbenchmarkingdiversepdf} for diverse document types and multi-granular capabilities, RealDoc-160 for robustness under challenging real-world capture conditions, and DocPTBench~\cite{du2025docptbench} for systematic evaluation of photographed document parsing.

\smallskip
\textbf{OmniDocBench}~\cite{ouyang2024omnidocbenchbenchmarkingdiversepdf} is a comprehensive and challenging benchmark designed to evaluate the performance and robustness of document parsing models across multiple dimensions. It encompasses diverse evaluation scenarios characterized by intricate layouts and rich content, including academic papers, textbooks, slides, research reports, and examination papers. To facilitate fine-grained structural analysis, the benchmark provides multi-granular annotations and evaluation metrics, covering layout analysis, text recognition, table parsing, formula extraction, and overall document structure understanding.

\smallskip
\textbf{RealDoc-160} is a self-constructed benchmark specifically designed to evaluate document parsing under authentic photographed conditions. It comprises 160 photographed document images: 80 English and 80 Chinese pages selected from in-house document collections. Each page was printed and then captured using mobile phones under diverse real-world conditions, including random lighting variations, arbitrary camera viewpoints, and natural paper deformations (\textit{e.g.}, bending, folding, creases). All OCR annotations were manually verified and corrected to ensure high-quality ground truth. This benchmark provides a rigorous testbed for assessing model robustness in practical document capture scenarios.

\smallskip
\textbf{DocPTBench}~\cite{du2025docptbench} is a comprehensive benchmark specifically designed for photographed document parsing and translation. Unlike prevailing benchmarks dominated by pristine scanned or digital-born documents, DocPTBench addresses the intricate challenges of real-world capture conditions, including geometric distortions and photometric variations. It comprises over 1,300 high-resolution photographed documents from multiple domains with meticulously human-verified annotations. Prior experiments on this benchmark reveal that transitioning from digital-born to photographed documents results in substantial performance decline, with specialized document parsing models showing an average accuracy decrease of 25\%, underscoring the unique challenges posed by real-world document capture.

\begin{table*}[htbp]
\setlength{\tabcolsep}{0.92mm}
  \small
  \caption{Comprehensive performance evaluations of document parsing on OmniDocBench~\cite{ouyang2024omnidocbenchbenchmarkingdiversepdf}. Y: Yes. N: No.} 
  \centering
  \begin{tabular}{clcc|cccccc}
    \toprule
    & \textbf{Methods} & \makecell[c]{\textbf{Single}\\\textbf{Model}} & \textbf{Size}  & \textbf{Overall}$\uparrow$ & \textbf{Text}$^{\text{Edit}}\downarrow$ & \textbf{Formula}$^{\text{CDM}}\uparrow$ & \textbf{Table}$^{\text{TEDS}}\uparrow$ & \textbf{Table}$^{\text{TEDS-S}}\uparrow$ & \textbf{Read Order}$^{\text{Edit}}\downarrow$ \\

    \midrule
    \multirow{3}{*}{\parbox[c]{2cm}{\centering\textbf{Pipeline}\\\textbf{Systems}}}
    & PP-StructureV3 & $N$ & - & 86.73 & 0.073 & 85.79 & 81.68 & 89.48 & 0.073 \\
    & Mineru2-pipeline & $N$ & - & 75.51 & 0.209 & 76.55 & 70.90 & 79.11 & 0.225 \\
    & Marker-1.8.2 & $N$ & - & 71.30 & 0.206 & 76.66 & 57.88 & 71.17 & 0.250 \\

    \midrule
    \multirow{6}{*}{\makecell[c]{\textbf{General}\\\textbf{VLMs}}}
    & Qwen3-VL-235B & $Y$ & 235B & 89.15 & 0.069 & 88.14 & 86.21 & 90.55 & 0.068 \\
    & Gemini-2.5 Pro & $-$ & - & 88.03 & 0.075 & 85.82 & 85.71 & 90.29 & 0.097 \\
    & Qwen2.5-VL & $Y$& 72B & 87.02 & 0.094 & 88.27 & 82.15 & 86.22 & 0.102 \\
    & InternVL3.5 & $Y$ & 241B & 82.67 & 0.142 & 87.23 & 75.00 & 81.28 & 0.125 \\
    & InternVL3 & $Y$ & 78B & 80.33 & 0.131 & 83.42 & 70.64 & 77.74 & 0.113 \\
    & GPT-4o & - & - & 75.02 & 0.217 & 79.70 & 67.07 & 76.09 & 0.148 \\
    
    \midrule
    \multirow{15}{*}{\makecell[c]{\textbf{Specialized}\\\textbf{VLMs}}}
    & PaddleOCR-VL & $N$ & 0.9B & \textbf{91.93} & \textbf{0.039} & \textbf{88.67} & \textbf{91.01} & \textbf{94.85} & \underline{0.048} \\
    & MinerU2.5 & $Y$ & 1.2B & \underline{90.67} & \underline{0.047} & \underline{88.46} & \underline{88.22} & \underline{92.38} & \textbf{0.044} \\
    & MonkeyOCR-pro & $N$ & 3B & 88.85 & 0.075 & 87.25 & 86.78 & 90.63 & 0.128 \\
    & dots.ocr & $Y$ & 3B & 88.41 & 0.048 & 83.22 & 86.78 & 90.62 & 0.053 \\
    & MonkeyOCR & $N$ & 3B & 87.13 & 0.075 & 87.45 & 81.39 & 85.92 & 0.129 \\
    & Deepseek-OCR & $Y$ & 3B & 87.01 & 0.073 & 83.37 & 84.97 & 88.80 & 0.086 \\
    & MonkeyOCR-pro & $N$ & 1.2B & 86.96 & 0.084 & 85.02 & 84.24 & 89.02 & 0.130 \\
    & Nanonets-OCR-s & $Y$ & 3B & 85.59 & 0.093 & 85.90 & 80.14 & 85.57 & 0.108 \\
    & MinerU2-VLM & $Y$ & 0.9B & 85.56 & 0.078 & 80.95 & 83.54 & 87.66 & 0.086 \\
    & olmOCR & $N$ & 7B & 81.79 & 0.096 & 86.04 & 68.92 & 74.77 & 0.121 \\
    & POINTS-Reader & $N$ & 3B & 80.98 & 0.134 & 79.20 & 77.13 & 81.66 & 0.145 \\
    & Mistral-OCR & - & - & 78.83 & 0.164 & 82.84 & 70.03 & 78.04 & 0.144 \\
    & OCRFlux & $N$ & 3B & 74.82 & 0.193 & 68.03 & 75.75 & 80.23 & 0.202 \\
    & Dolphin & $Y$ & 0.3B & 74.67 & 0.125 & 67.85 & 68.70 & 77.77 & 0.124 \\
    & Dolphin-v1.5 & $Y$ & 0.3B & 85.06 & 0.085 & 79.44	& 84.25 & 88.06 & 0.071 \\
    & \textbf{Dolphin-v2 (Ours)} & $Y$ & 3B & 89.78 & 0.054 & 87.63 & 87.02 & 90.48 & 0.054
    \\
    \bottomrule
  \end{tabular}\label{tab:omni_res}
\end{table*}

\section{Experiments}
\label{sec:exps}
\subsection{Implementation Details}
We fine-tune Qwen2.5-VL-3B~\cite{bai2025qwen2} as our backbone. During training, the visual encoder parameters are frozen while both the vision-language adapter and language model decoder are trained end-to-end. We employ the AdamW optimizer with an initial learning rate of $8 \times 10^{-5}$, weight decay of 0, and cosine annealing schedule with a warmup ratio of 0.03. The model is trained for 10 epochs on 40 A100 GPUs with a per-device batch size of 8 and gradient accumulation over 4 steps, yielding an effective batch size of 32 per device. The maximum sequence length is set to 131,072 tokens to accommodate long documents and complex layouts.

\subsection{Comparison with Existing Methods}
Comprehensive evaluations are conducted on the above datasets. We next discuss the results.

\smallskip
\textbf{Qualitative Analysis}.
Figure~\ref{fig:qualitative} presents representative parsing results of Dolphin-v2 on photographed documents with various distortions. Unlike digital documents with clean layouts, photographed documents often exhibit challenging characteristics such as perspective transformations, surface wrinkles, and non-uniform illumination. As shown in the examples, Dolphin-v2 demonstrates robust performance in handling these distortions while maintaining accurate text recognition and structure preservation across diverse document types. These results highlight the effectiveness of our holistic parsing strategy for photographed documents, which processes the entire page as a unified context rather than fragmenting it into isolated elements.

Figure~\ref{fig:code_qualitative} demonstrates the capability of Dolphin-v2 in preserving code structure. Unlike PaddleOCR-VL~\cite{cui2025paddleocrvl}, which fails to maintain indentation, Dolphin-v2 accurately preserves the hierarchical structure of code through dedicated code parsing with the specialized prompt $P_\text{code}$, ensuring syntactic correctness essential for programming languages.
This improvement stems from the explicit separation of code blocks as a distinct element category during layout analysis, which provides clear semantic priors that guide the decoder to attend to whitespace patterns and indentation levels that would otherwise be ignored in general text parsing. This design choice also highlights the extensibility of our framework: new element types can be incorporated by simply introducing additional layout categories and corresponding prompts.

\begin{table}[t]
  \small
  \caption{Performance comparison on RealDoc-160 benchmark. All scores are Edit Distance (lower is better).}
  \centering
  \resizebox{1\linewidth}{!}{
  \begin{tabular}{l l c | c c c}
    \toprule
    \textbf{Type} & \textbf{Methods} & \textbf{Size} & \textbf{EN}$\downarrow$ & \textbf{ZH}$\downarrow$ & \textbf{AVG}$\downarrow$ \\
    \midrule
    \multirow{4}{*}{\makecell[l]{\textbf{General}\\\textbf{VLMs}}}
    & Gemini-2.5 Pro & -  & 0.0889 & \textbf{0.0681} & 0.0785 \\
    & GPT-4o         & -  & 0.0588 & 0.2694 & 0.1641 \\
    & GPT-4o-mini    & -  & 0.1943 & 0.7057 & 0.4500 \\
    & Qwen2.5-VL     & 7B & 0.0425 & 0.1144 & 0.0785 \\
    \midrule
    \multirow{7}{*}{\makecell[l]{\textbf{Specialized}\\\textbf{VLMs}}}
    & Dolphin      & 0.3B & 0.3867             & 0.4859 & 0.4363 \\
    & MonkeyOCR    & 3B   & \underline{0.0362} & 0.1180 & \underline{0.0771} \\
    & MinerU2.5    & 1.2B & 0.3359             & 0.3641 & 0.3500 \\
    & PaddleOCR-VL & 0.9B & 0.1616             & 0.1979 & 0.1798 \\
    & Dolphin-1.5  & 0.3B & 0.1847             & 0.2614 & 0.2231 \\
    & dots.ocr     & 3B   & 0.1480             & 0.2553 & 0.2017 \\
    & Dolphin-v2   & 3B   & \textbf{0.0046}    & \underline{0.0737} & \textbf{0.0392} \\
    \bottomrule
  \end{tabular}
  }
  \label{tab:performance_comparison}
\end{table}

\begin{table*}[htbp]
\centering
\caption{Performance evaluation of document parsing models on photographed documents from DocPTBench~\cite{du2025docptbench}.}
\label{tab:parsing_results_photo}
\begin{tabular*}{\textwidth}{@{\extracolsep{\fill}} l l ll ll ll ll ll @{}}
\toprule
\multirow{2}{*}{\textbf{Type}} & \multirow{2}{*}{\textbf{Methods}} & \multicolumn{2}{c}{\textbf{Overall$^{\mathbf{Edit}\downarrow}$}} & \multicolumn{2}{c}{\textbf{Text$^{\mathbf{Edit}\downarrow}$}} & \multicolumn{2}{c}{\textbf{Formula$^{\mathbf{Edit}\downarrow}$}} & \multicolumn{2}{c}{\textbf{Table$^{\mathbf{TEDS}\uparrow}$}} & \multicolumn{2}{c}{\textbf{Read Order$^{\mathbf{Edit}\downarrow}$}} \\
\cmidrule(lr){3-4} \cmidrule(lr){5-6} \cmidrule(lr){7-8} \cmidrule(lr){9-10} \cmidrule(lr){11-12}
& & \textbf{En} & \textbf{Zh} & \textbf{En} & \textbf{Zh} & \textbf{En} & \textbf{Zh} & \textbf{En} & \textbf{Zh} & \textbf{En} & \textbf{Zh} \\
\midrule
\multirow{6}{*}{\textbf{General VLMs}}
& Qwen2.5-VL-72B~\cite{bai2025qwen2} & 41.5 & 57.0 & 36.2 & 56.6 & 42.2 & 61.8 & 57.0 & 55.5 & 28.1 & 51.3 \\
& Gemini2.5-Pro~\cite{comanici2025gemini25pushingfrontier} & \textbf{18.2} & \textbf{30.4} & \textbf{9.8} & \textbf{22.0} & \textbf{37.1} & \textbf{52.2} & \textbf{81.3} & \textbf{82.9} & \textbf{11.2} & \textbf{18.1} \\
& Doubao-1.6-v~\cite{guo2025seed1} & 54.7 & 55.4 & 60.6 & 58.2 & 51.5 & 61.1 & 27.6 & 37.9 & 39.7 & 40.2 \\
& Qwen-VL-Max~\cite{Qwen-VL} & 27.7 & 42.7 & 15.9 & 41.5 & 41.8 & 57.2 & 71.1 & 71.6 & 16.8 & 34.4 \\
& GLM-4.5v~\cite{vteam2025glm45vglm41vthinkingversatilemultimodal} & 36.7 & 49.6 & 26.2 & 47.7 & 49.9 & 66.2 & 58.9 & 54.0 & 27.3 & 35.7 \\
& Kimi-VL~\cite{team2025kimi} & 36.5 & 38.7 & 17.2 & 22.0 & 48.6 & 52.2 & 57.1 & 67.8 & 14.3 & 18.1 \\
\midrule
\multirow{12}{*}{\textbf{Specialized VLMs}}
& PaddleOCR-VL~\cite{cui2025paddleocr} & 37.5 & 39.6 & 29.4 & 37.7 & 46.5 & 52.6 & 54.2 & 65.3 & 28.8 & 37.9 \\
& MinerU2.5~\cite{niu2025mineru25decoupledvisionlanguagemodel} & 37.3 & 47.4 & 37.0 & 53.6 & 44.3 & 62.0 & 54.9 & 59.8 & 29.0 & 40.3 \\
& dots.ocr~\cite{dots_ocr} & 33.7 & 37.3 & 29.8 & 35.8 & \textbf{39.2} & 54.4 & 63.7 & \textbf{67.6} & 32.8 & 31.8 \\
& MonkeyOCR~\cite{li2025monkeyocr} & 46.4 & 52.8 & 34.5 & 43.9 & 48.7 & 61.6 & 33.1 & 37.4 & 37.9 & 44.1 \\
& Dolphin~\cite{feng2025dolphin} & 57.5 & 71.5 & 54.9 & 71.5 & 65.6 & 82.8 & 33.0 & 19.3 & 46.2 & 57.7 \\
& olmOCR~\cite{poznanski2025olmocr} & 39.1 & 46.1 & 19.3 & 27.2 & 50.7 & 66.9 & 56.5 & 56.9 & 20.7 & 24.4 \\
& OCRFlux~\cite{OCRFlux2025} & 36.2 & 45.8 & 30.4 & 40.4 & 48.4 & 81.1 & 49.5 & 54.3 & 22.5 & 32.1 \\
& SmolDocling~\cite{nassar2025smoldocling} & 90.1 & 93.7 & 89.8 & 99.2 & 99.6 & 99.9 & 4.4 & 2.4 & 72.7 & 75.9 \\
& Nanonets-OCR~\cite{Nanonets-OCR-S} & 38.6 & 52.1 & 21.0 & 42.0 & 48.1 & 67.0 & 58.5 & 50.6 & 21.4 & 32.7 \\
& DeepSeek-OCR~\cite{wei2025deepseek} & 54.4 & 57.8 & 56.7 & 57.6 & 54.4 & 74.1 & 28.0 & 35.4 & 41.7 & 40.4 \\
& Nanonets-OCR2~\cite{Nanonets-OCR-S} & 34.2 & 46.1 & 25.5 & 44.6 & 69.0 & 76.4 & \textbf{70.7} & 66.0 & 19.5 & 31.4 \\
& \textbf{Dolphin-v2 (Ours)} & \textbf{30.8} & \textbf{37.3} & \textbf{21.5} & \textbf{35.8} & 48.2 & \textbf{54.4} & 59.0 & 53.6 & \textbf{19.0} & \textbf{29.2} \\
\bottomrule
\end{tabular*}
\end{table*}

Figure~\ref{fig:qualitative_case2} showcases the complete two-stage parsing pipeline of Dolphin-v2 on a complex academic document containing dense formulas and reference sections. The visualization illustrates the entire workflow: from the input document image (left), through layout analysis with predicted element boundaries and reading order (middle), to the final rendered Markdown output (right). Dolphin-v2 accurately handles challenging scenarios, including extensive mathematical expressions and bibliography entries, demonstrating its robustness across diverse document structures. The success in such challenging layouts can be attributed to our anchor-based parallel parsing strategy, where the first-stage layout predictions decompose the complex page into manageable elements, allowing the second stage to focus on content extraction without being overwhelmed by the global structural complexity.

\smallskip
\textbf{Quantitative Results}.
Table~\ref{tab:omni_res} presents a comprehensive comparison of Dolphin-v2 against state-of-the-art methods on OmniDocBench (v1.5). Dolphin-v2 achieves an overall score of 89.45, significantly outperforming the original Dolphin (74.67) by +14.78 points. Moreover, as a single-model solution with 3B parameters, it demonstrates competitive performance compared to both specialized OCR models and large-scale general VLMs, such as Qwen2.5-VL (72B, 87.02). Dolphin-v2 achieves balanced performance across all metrics, including text recognition (0.054 Edit distance), reading-order prediction (0.054 Edit distance), and table structure parsing (90.48 TEDS-S).

\begin{table}[t]
\centering
\caption{Ablation study on document type classification. Results are evaluated on the RealDoc-160 benchmark with Edit Distance for text.}
\label{tab:ablation_classification}
\small
\begin{tabular}{lccc}
\toprule
\textbf{Model Variant} & \textbf{EN}$\downarrow$ & \textbf{ZH}$\downarrow$ & \textbf{AVG}$\downarrow$ \\
\midrule
w/o Classification & 0.1523 & 0.2218 & 0.1871  \\
Dolphin-v2 (Full)  & \textbf{0.0046} & \textbf{0.0737} & \textbf{0.0392} \\
\bottomrule
\end{tabular}
\end{table}

Table~\ref{tab:performance_comparison} presents the performance comparison on our self-constructed RealDoc-160 benchmark, which specifically evaluates model robustness under real-world photographed conditions. Dolphin-v2 achieves the best average performance with an Edit Distance of 0.0392, significantly outperforming both general VLMs and specialized OCR models. Notably, Dolphin-v2 demonstrates exceptional performance on English documents (0.0046), surpassing the second-best MonkeyOCR (0.0362) by 87.3\%. For Chinese documents, Dolphin-v2 achieves 0.0737, ranking second only to Gemini-2.5 Pro (0.0681). These results validate the effectiveness of our holistic parsing strategy for photographed documents with realistic distortions.

Table~\ref{tab:parsing_results_photo} presents the evaluation results on DocPTBench for photographed document parsing. Among specialized VLMs, Dolphin-v2 achieves the best overall performance with Edit distances of 30.8 (En) and 37.3 (Zh), substantially improving over the original Dolphin (57.5 and 71.5). Dolphin-v2 also attains the lowest Edit distances in text recognition and reading-order prediction among specialized models. While Gemini2.5-Pro achieves the best overall performance, Dolphin-v2 narrows the gap considerably as a 3B model, particularly excelling in reading-order prediction (19.0 En, 29.2 Zh). These results further validate the effectiveness of our hybrid parsing strategy for photographed documents.

\begin{table}[t]
\centering
\caption{Ablation study on formula parsing strategy on OmniDocBench~\cite{ouyang2024omnidocbenchbenchmarkingdiversepdf}.}
\label{tab:ablation_formula}
\small
\begin{tabular}{lc}
\toprule
\textbf{Model Variant} & \textbf{CDM}$\uparrow$  \\
\midrule
Unified Parsing & 83.34  \\
Dedicated Formula Parsing & \textbf{86.72} \\
\bottomrule
\end{tabular}
\end{table}

\subsection{Ablation Studies}
We conduct extensive experiments to validate the
effectiveness of the core strategies in Dolphin-v2.

\smallskip
\textbf{Document Type Classification.} 
To validate the necessity of document type classification in the first stage, we compare two model variants: one trained with document type classification and one without classification. 
Table~\ref{tab:ablation_classification} presents the results on the RealDoc-160 benchmark, which contains diverse photographed documents captured under authentic real-world conditions with various distortions, lighting variations, and viewpoint changes. 
The results demonstrate that without document type classification, the model exhibits significantly degraded performance, with the average Edit Distance increasing from 0.0392 to 0.1871 (a 377\% degradation). 
Specifically, the performance drop is substantial for both English (0.0046 vs. 0.1523) and Chinese (0.0737 vs. 0.2218) documents.
This is because a unified parsing strategy cannot optimally handle both document types: directly applying holistic parsing to digital documents sacrifices the benefits of element-wise parallel processing and type-specific prompting, while applying element-wise parsing to photographed documents fails to handle distortions and irregular layouts. 
The classification-based hybrid strategy enables Dolphin-v2 to adaptively select the optimal parsing approach for different documents, thereby achieving superior robustness across diverse real-world scenarios.

\begin{figure}[t]
  \centering
  \includegraphics[width=1\columnwidth]{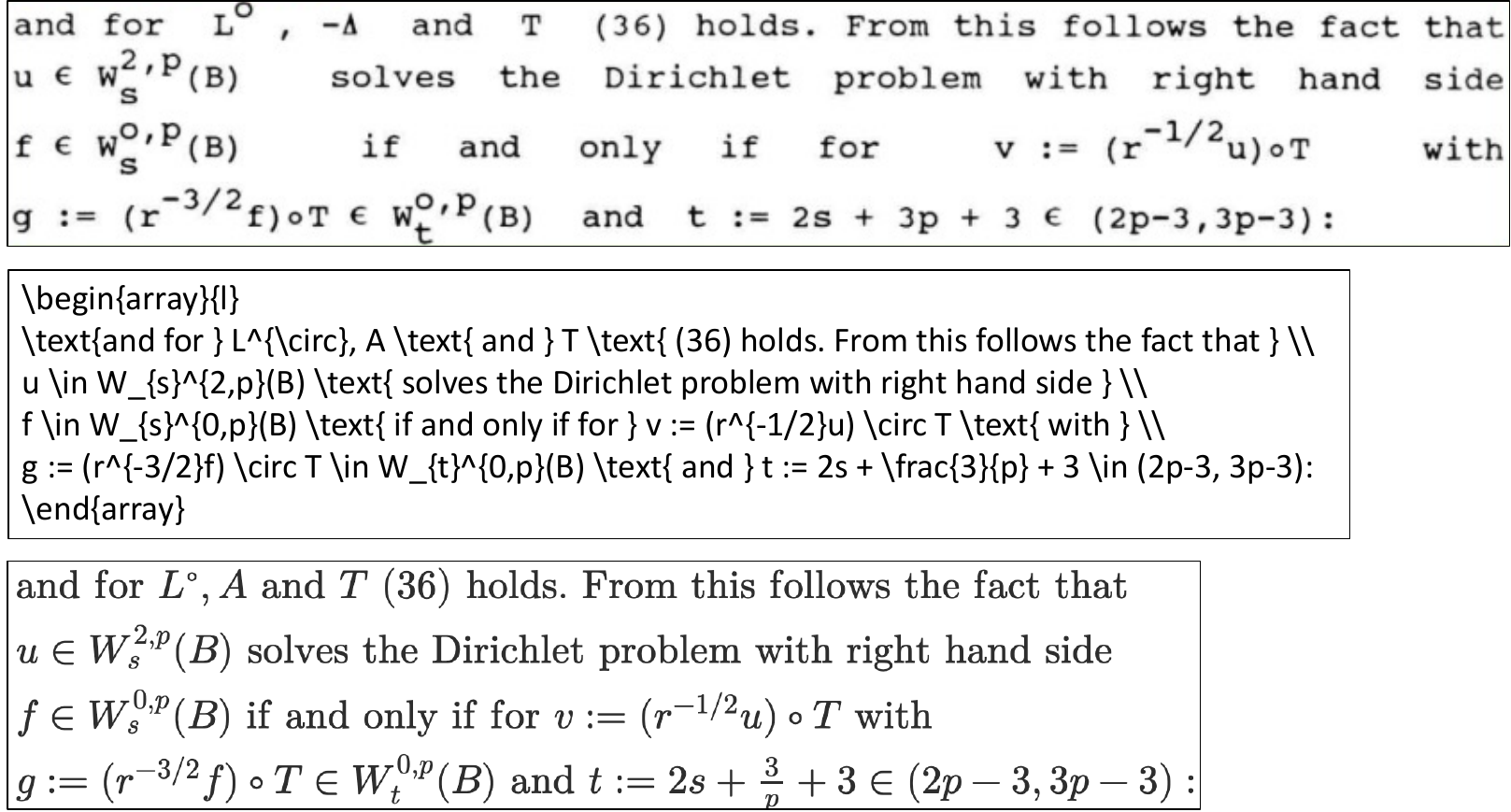}
\caption{Illustration of formula parsing confusion without dedicated formula handling. \textbf{Top}: the input formula image. \textbf{Middle}: the parsing result from the Dolphin-v2 variant without text block/formula separation. \textbf{Bottom}: the rendered image based on the parsing result. Although the rendered output appears visually similar to the input, the model incorrectly treats the entire text block as a standalone display formula during parsing.}
      \label{fig:formula_bad_case}
\end{figure}

\smallskip
\textbf{Formula Parsing Strategy.} 
We investigate the impact of dedicated formula parsing by comparing two approaches: (1) parsing formulas separately with a specialized prompt $P_\text{formula}$, and (2) treating formulas as regular text within paragraph blocks following the original Dolphin~\cite{feng2025dolphin}. 
Table~\ref{tab:ablation_formula} shows the results on the formula subset of OmniDocBench. 
Without dedicated formula parsing, the model may misclassify text blocks containing mathematical expressions as standalone formulas. 
This confusion arises from ambiguous contextual cues when formulas and text share the same processing pipeline. 
As illustrated in Figure~\ref{fig:formula_bad_case}, although the rendered output appears visually similar to the input, the model without dedicated formula parsing fundamentally treats the text block as a standalone formula during parsing, demonstrating the necessity of explicit text/formula decomposition.
By introducing separate formula detection in layout analysis and specialized LaTeX generation prompts, Dolphin-v2 achieves an improvement of 3.38 points in formula parsing accuracy (\textit{i.e.}, CDM: 86.72 vs. 83.34). 
This enhancement addresses a key limitation of the original Dolphin, which suffered from formula-text confusion issues due to the lack of dedicated formula handling.

\section{Limitation Discussions}
\label{sec:limitation}
\begin{figure}[t]
\centering
\includegraphics[width=1\columnwidth]{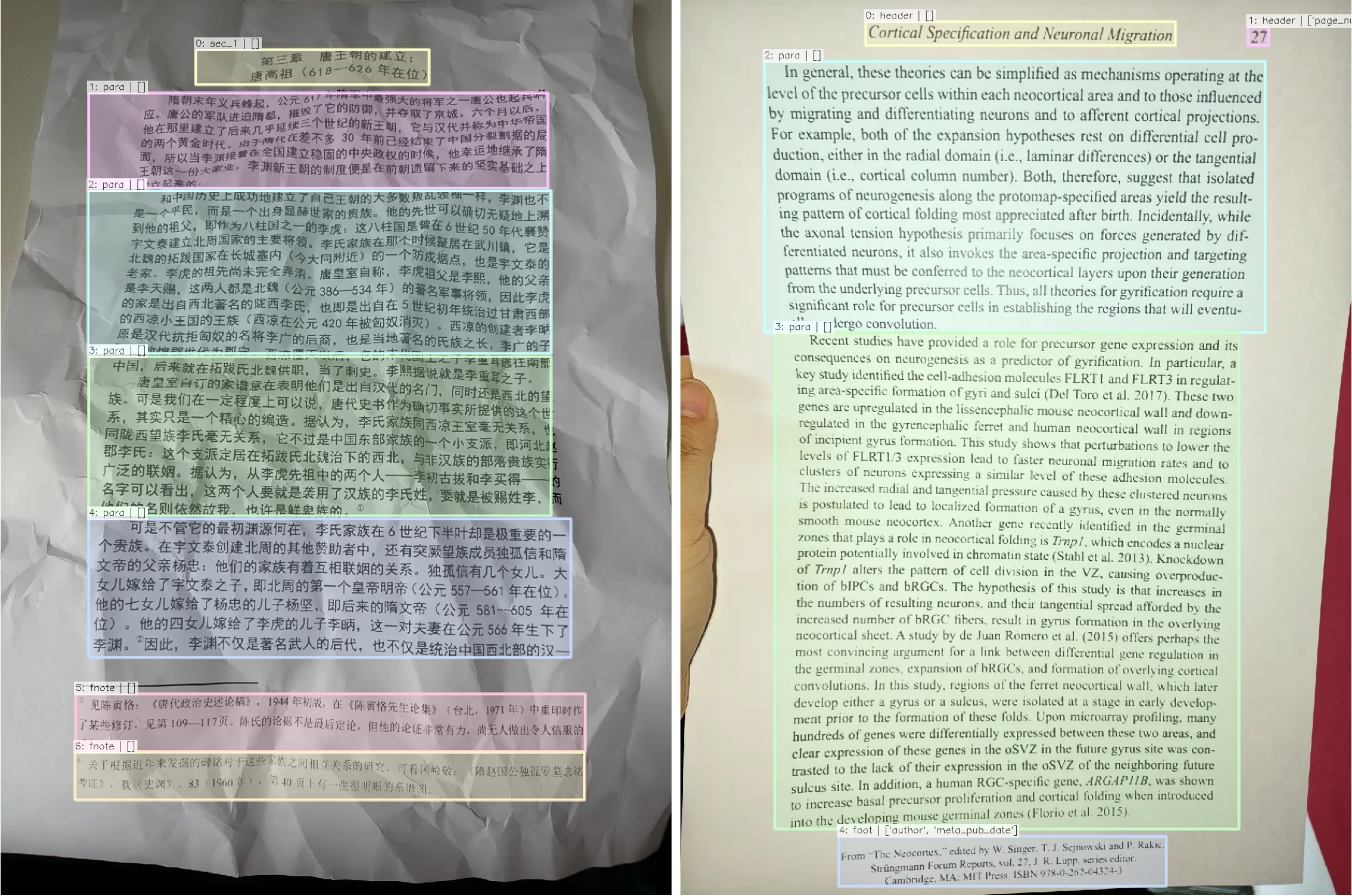}
\caption{Failure cases of document type classification in Stage 1. These photographed documents with mild distortions are incorrectly classified as digital documents, leading to element-wise layout parsing.}
\label{fig:bad_case}
\end{figure}

While Dolphin-v2 demonstrates strong performance across diverse document scenarios, we acknowledge several limitations that warrant future investigation, which are detailed as follows:

\smallskip
\textbf{Document Type Classification Errors.}
While our joint classification and layout analysis stage effectively distinguishes between digital-born and photographed documents in most cases, misclassification occasionally occurs in borderline scenarios. As illustrated in Figure~\ref{fig:bad_case}, photographed documents with mild distortions may be incorrectly classified as digital documents, leading to inappropriate element-wise parsing rather than holistic page-level parsing. Such errors typically arise when the distortions are subtle, for instance in documents captured at near-perpendicular angles with minimal wrinkles or lighting variations. This misclassification can propagate to the second stage, potentially degrading parsing quality. Future work could explore more robust classification strategies to better handle these ambiguous cases.

\smallskip
\textbf{Element Type Coverage.}
Currently, Dolphin-v2 supports the parsing of text paragraphs, formulas, tables, and code blocks. While this coverage addresses the majority of document parsing scenarios, certain specialized element types remain unexplored. For instance, chemical structure formulas (e.g., organic chemistry diagrams), charts and data visualizations, and musical notations represent valuable targets for future extension. Note that our scalable anchor-based framework naturally accommodates such extensions. Adding new element types requires only introducing corresponding category labels in the layout analysis stage and designing type-specific prompts for the content parsing stage, without necessitating fundamental architectural changes. This scalability positions Dolphin-v2 as a flexible foundation for comprehensive document understanding systems that can evolve with emerging document types and user requirements.

\section{Conclusion}
\label{sec:conclusion}
In this paper, we present Dolphin-v2, a universal document parsing model that seamlessly handles diverse document types through a two-stage framework. By jointly performing document type classification and layout analysis in the first stage, followed by a hybrid parsing strategy that processes photographed documents holistically while parsing digital documents element-wise in parallel, Dolphin-v2 achieves both high accuracy and computational efficiency. Compared to the original Dolphin, Dolphin-v2 introduces three key enhancements: (1) finer-grained element detection with 21 categories, reading-order prediction, and semantic attribute extraction, (2) absolute coordinate representation for precise spatial localization, and (3) dedicated parsing modules for formulas and code blocks with indentation preservation. Extensive experiments validate the effectiveness of our approach, achieving +14.78 points improvement on OmniDocBench and 91\% error reduction on photographed documents, demonstrating that our unified architecture successfully bridges the gap between specialized and general document parsing capabilities.

\smallskip
\textbf{Broad Impact.}
Moreover, our Dolphin-v2 has the potential to benefit a wide range of real-world applications. By enabling accurate and efficient parsing of diverse document types, our model can facilitate document digitization in libraries and archives, improve accessibility for visually impaired users through reliable text extraction, and support knowledge management in enterprise settings. The ability to handle photographed documents with distortions is particularly valuable in resource-limited environments where high-quality scanning equipment is unavailable.

\smallskip
\textbf{Future Work.}
Building upon the findings and identified limitations of this work, several promising directions are proposed for future exploration, as detailed below:

\begin{itemize} \item \textbf{Enhancing Classification Robustness:} To mitigate misclassification in borderline scenarios, we aim to refine the document type discriminator. Future efforts will focus on integrating more nuanced geometric and photometric features to better distinguish between near-perpendicular photographed documents and digital-born counterparts.

\item \textbf{Expanding Element Coverage:} Leveraging the inherent scalability of the anchor-based framework, we plan to augment the system's capability to parse more specialized elements, including chemical structures, complex charts, and musical notations, by incorporating category-specific prompts and data.

\item \textbf{Cross-page Context Modeling:} We intend to extend the framework from single-page analysis to multi-page document understanding. This involves developing cross-page attention mechanisms or memory modules to maintain consistency for elements that span across page boundaries.

\item \textbf{Downstream Integration:} Finally, we aim to integrate Dolphin-v2 with advanced Large Language Models (LLMs) to facilitate end-to-end document intelligence tasks, such as complex reasoning, information extraction, and document-based question answering.

\end{itemize}

\ifCLASSOPTIONcaptionsoff
  \newpage
\fi

{
\bibliographystyle{IEEEtran}
\bibliography{IEEEabrv,egbib}
}

\end{document}